\documentclass{article} 
\usepackage{iclr2021_conference,times}

\usepackage{amsmath,amsfonts,bm}

\def\eqref#1{equation~\ref{#1}}

\def\1{\bm{1}}

\DeclareMathAlphabet{\mathsfit}{\encodingdefault}{\sfdefault}{m}{sl}
\SetMathAlphabet{\mathsfit}{bold}{\encodingdefault}{\sfdefault}{bx}{n}

\usepackage{hyperref}
\usepackage{url}
\usepackage{amsmath, amssymb, amsthm}
\usepackage{mathrsfs}
\usepackage{geometry}
\usepackage{extarrows}
\usepackage{xcolor}
\usepackage{graphicx}
\usepackage{booktabs}
\usepackage{multicol}
\usepackage{multirow}
\usepackage{float}

\def\cL{\mathcal{R}}

\def\bx{\boldsymbol{x}}

\def\by{\boldsymbol{y}}

\def\bm{\boldsymbol{m}}
\usepackage{amsmath}

\def\cL{\mathcal{L}}

\title{Feature Protection For Out-of-distribution Generalization}

\author{Antiquus S.~Hippocampus, Natalia Cerebro \& Amelie P. Amygdale \thanks{ Use footnote for providing further information
about author (webpage, alternative address)---\emph{not} for acknowledging
funding agencies.  Funding acknowledgements go at the end of the paper.} \\
Department of Computer Science\\
Cranberry-Lemon University\\
Pittsburgh, PA 15213, USA \\
\texttt{\{hippo,brain,jen\}@cs.cranberry-lemon.edu} \\
\And
Ji Q. Ren \& Yevgeny LeNet \\
Department of Computational Neuroscience \\
University of the Witwatersrand \\
Joburg, South Africa \\
\texttt{\{robot,net\}@wits.ac.za} \\
\AND
Coauthor \\
Affiliation \\
Address \\
\texttt{email}
}

\begin{document}

\maketitle

\begin{abstract}
With the availability of large pre-trained models, a modern workflow for building real-world machine learning solutions is to fine-tune such models on a downstream task with a relatively small domain-specific dataset. 
In such applications, one major challenge is that the small fine-tuning dataset does not have sufficient coverage of the distribution encountered when the model is deployed. It is thus important to design fine-tuning methods that are robust to out-of-distribution (OOD) data that are under-represented by the training data. 
This paper compares common fine-tuning methods to investigate their OOD performance and demonstrates that standard methods will result in a significant change to the pre-trained model so that the fine-tuned features overfit the fine-tuning dataset. However, this causes deteriorated OOD performance. 
To overcome this issue, we show that protecting pre-trained features leads to a fine-tuned model more robust to OOD generalization. 
We validate the feature protection methods with extensive experiments of fine-tuning CLIP on ImageNet and DomainNet.
\end{abstract}

\section{Introduction}
It is a common practice to pre-train a model on a large-scale dataset and then perform fine-tuning on a downstream task with limited samples. The model fine-tuned on the extra application-specific data often yields large performance gains in the fine-tuned domain. 
The fine-tuning dataset is often small and does not have sufficient coverage of the distribution encountered. As a result, recent works show the fine-tuned model deteriorates dramatically when applied to a test distribution, which is different from the fine-tuning distribution (\cite{wortsman2022robust,kumar2022fine}). In other words, the fine-tuning procedure can lead to better in-distribution (ID) performance but worse out-of-distribution (OOD) performance.

\begin{figure}[t]
    \centering
    \includegraphics[scale=0.5]{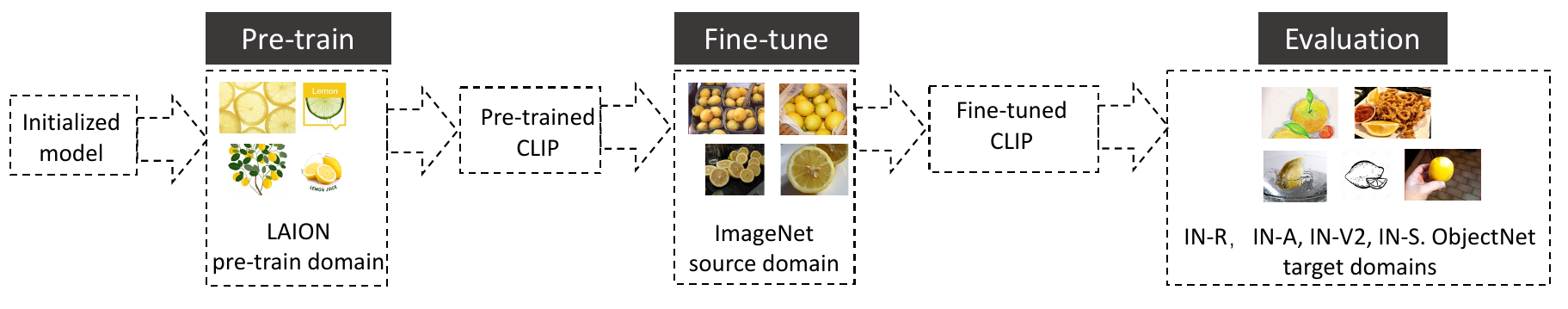}
    \caption{Illustration of the workflow: Pre-train, fine-tune, and evaluation. Nowadays, large neural networks are often pre-trained on large-scale datasets (e.g., LAION). Then, the pre-trained model is fine-tuned on a small dataset from a fine-tuning (source) domain(e.g., ImageNet). Finally, we evaluate the model on target domains. For example, we fine-tune the pre-trained CLIP model on the ImageNet dataset with photo style, but the testing samples are from other styles such as sketch, art, cartoon, etc.}
    \label{fig:three_domains}
\end{figure}

 We provide an in-depth feature-level explanation for the OOD performance drop of the fine-tuned model. 
 Figure \ref{fig:three_domains} illustrates the pre-training and fine-tuning workflow and possible distributional shifts. 
 The observations in Section~\ref{sect:investigation} show a  significant change of the pre-trained model during fine-tuning to fit the fine-tuning domain, causing the OOD performance to drop. 
These observations indicate the necessity to protect pre-trained features.

We conduct a systematic investigation of various methods developed across different communities to preserve pre-trained features, including (1) Continual learning methods such as L1, L2, and Knowledge Distillation; (2) Parameter efficient fine-tuning such as LoRA (\cite{hu2021lora}); (3) WiSE-FT, a promising model averaging method in OOD community; 
Our findings indicate that (1) continual learning methods such as L1, L2, and KD penalty can effectively protect pre-trained features compared to vanilla fine-tuning while still achieving reasonable performance on the fine-tuned task. (2) LoRA excels in fine-tuning tasks compared to other tasks.  (3) The model averaging method in the OOD community, WiSE-FT, demonstrates the strongest performance.

\section{An investigation on why fine-tuning hurts OOD performance}
\label{sect:investigation}
\textbf{Notations.}
Let random variables $\bx$ and $\by$ denote the input and output, respectively. 
Our goal is to learn a function $f_\theta$ to predict $\by$ based on $\bx$, where $\theta$ is the parameter. We consider $f$ composed of a feature extractor $\Phi$ and a classifier $v$, i.e., $f = [\Phi, v]$.  A pre-trained model can be noted as $f_{\theta_{0}}$.

Previous work(\cite{andreassen2021evolution}) shows that OOD performance drops on the convergence of fine-tuning. We also have a similar observation. After fine-tuning pre-trained CLIP on ImageNet, the ID accuracy of ImageNet is improved from 68.33 to 81.47. But the average OOD accuracy on 5 variants of ImageNet datasets drops from 58.32 to 53.83. 
However, it is still uncertain whether this discrepancy arises due to the feature extractor or classifier head.
Recent researchers hold the view that the pre-trained feature extractor is good enough for both ID and OOD tasks. It is the sub-optimal classifier that caused the decline of OOD results. For instance,  \cite{rosenfeld2022domain, qiu2023simple, kirichenko2022last} suggest that $\Phi$ still encodes features as effectively as $\Phi_0$ and the OOD performance decline since $v$ is sub-optimal for the target domain. 

\begin{figure}[t]
    \centering
    \includegraphics[width=0.8\linewidth]{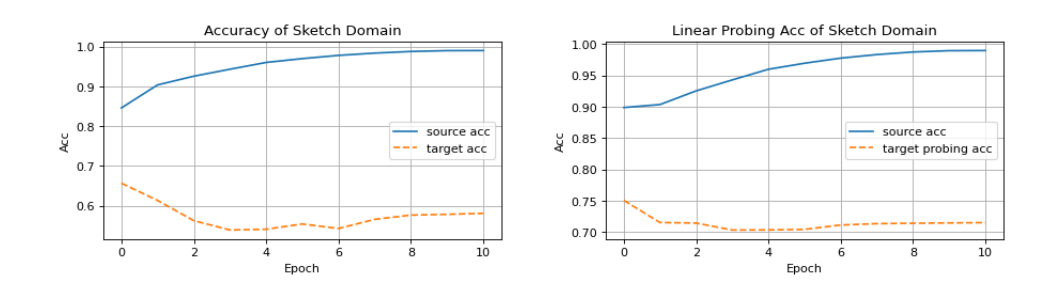}
    \caption{Accuracy and linear probing accuracy of sketch domain in DomainNet.}
    \label{fig:acc_n_linear_prob_acc}
\end{figure}

To isolate the effect of the classifier discussed above, we perform the following experiments to show that the OOD decline is closely related to distortion of semantic representation(\cite{davari2022probing}). Specifically, we store a sequence of checkpoints, namely, $[\Phi_1, v_1], [\Phi_2, v_2], ..., [\Phi_t, v_t]$ during fine-tuning on the source domain. We perform linear probing for each checkpoint on the target domain respectively, i.e., obtain $\bar{v}_i=\underset{v}{\arg \min } \mathcal{L}_t\left(\left[\Phi_i, v\right]\right)$, where $\cL_{t}([\Phi_i, v])$ is the loss on the target domain of $[\Phi_i, v]$. Now that we have the $\bar v_i$, which is optimal for the target domain given the feature $\Phi_i$. We name the accuracy of $[\Phi_i, \bar v_i]$ on the target domain as \textit{linear probing accuracy}. It measures the effectiveness of the feature representation by $\Phi_i$ for the target domain. In Figure~\ref{fig:acc_n_linear_prob_acc}, we show the trend of accuracy  $[\Phi_i, v_i]$ as well as linear probing accuracy $[\Phi_i, \bar v_i]$ on sketch target domain. Refer to Appendix~\ref{app:all_target_domains} for all target domains. The linear probing accuracy also drops during fine-tuning, indicating that the fine-tuning process indeed distorts semantic features.

\section{Preserving pre-trained feature}
\vspace{-0.32cm}
\subsection{Datasets and models}
Following \cite{Wortsman2021wiseft, Goyal2022FLYP, Tian2023TrainablePG, Lin2023MultimodalityHU}, we conduct experiments on pre-trained CLIP ViT-B/16(\cite{radford2021learning}) with the following settings:
\textbf{(1)ImageNet Settings.}
Fine-tune CLIP on ImageNet and evaluate on 5 variants of ImageNet with distributional shifts(ImageNet-V2(\cite{recht2019imagenet}), ImageNet-R(\cite{hendrycks2021many}), ImageNet Sketch(\cite{wang2019learning}), ObjectNet(\cite{barbu2019objectnet}), and ImageNet-A(\cite{hendrycks2021natural}). 
\textbf{(2) DomainNet Settings.}
Fine-tune CLIP on the ``real" domain of DomainNet(\cite{peng2019moment}) and assess the extent of catastrophic forgetting on the rest domains of DomainNet(``Clipart", ``Infograph", ``Painting", ``Quickdraw", and ``Sketch").
Illustration samples of these datasets are shown in Figure~\ref{fig:imagenet_samples}
We include more details on the experiments in Appendix~\ref{app:exp_details}. 

\subsection{Methods}
To validate the hypothesis that preserving the pre-trained features can benefit the OOD performance of fine-tuned models, we examine existing methods developed across different communities to preserve pre-trained features. Though these methods are not originally designed for OOD tasks, we anticipate that they can help OOD generalization by preserving more robust features. 

\textbf{Regularization towards pre-trained weight.}
Let's recall that $\theta_0$ represents the parameters of the pre-trained foundation model. To protect the pre-trained feature, a straightforward approach is to enforce a constraint on the proximity of $\theta$ to $\theta_0$. In other words, we ensure that $\theta$ does not deviate too far from $\theta_0$ (\cite{xuhong2018explicit}). We accomplish this by optimizing two penalties:
(1) The L1 penalty $|\theta - \theta_0|$ (\cite{panigrahi2023task})\footnote{\cite{panigrahi2023task} applies a post-processing technique to find the sparsity structure in the $\theta - \theta_0$. For simplicity, we use the L1 norm to encourage sparsity (\cite{zhao2006model}). This method is also connected with parameter-efficient fine-tuning.};
(2) The L2 penalty $\|\theta - \theta_0\|_2^2$ (\cite{xuhong2018explicit}).

\textbf{Parameter-efficient Fine-tuning}
Parameter-efficient fine-tuning aims to achieve comparable performance as traditional fine-tuning while utilizing significantly fewer trainable parameters. One widely adopted method in this domain is LoRA (Low-Rank Adaptation)(\cite{Hu2021LoRALA}), which effectively represents modified weights $\Delta\theta$ using low-rank matrix pairs while keeping most of the pre-trained network parameters frozen. For more details, refer to Appendix~\ref{app:LoRA_details}.

\textbf{Knowledge Distillation}
Knowledge distillation involves transferring knowledge from a larger model (teacher) to a smaller one (student). 
In our case, we aim to preserve the features of the pre-trained model during the fine-tuning process. 
We utilize the pre-trained model $f_{\theta_0}$ as the teacher and the fine-tuned model $f_{\theta}$ as the student. 
To ensure the student model's predictions or learned features align closely with those of the teacher model, we enforce an L2 regularization constraint on their outputs: $\| f_{\theta}(\mathbf{x}) - f_{\theta_0}(\mathbf{x})\|_2^2$ (\cite{buzzega2020dark,huang2021continual}).

\textbf{Model Averaging}
The model averaging method, WiSE-FT, introduced in \cite{Wortsman2021wiseft}, suggests a linear interpolation approach between the pre-trained parameter $\theta_0$ and the fine-tuned parameter $\theta$. 
This results in the model $f_{(1-\alpha) \theta_0 + \alpha \theta}$, where $\alpha$ represents a hyper-parameter ranging from 0 to 1. 
It is a popular method from the OOD community with strong performance.

\section{Results}

\begin{figure}[t]
    \centering
    \includegraphics[width=0.45\linewidth]{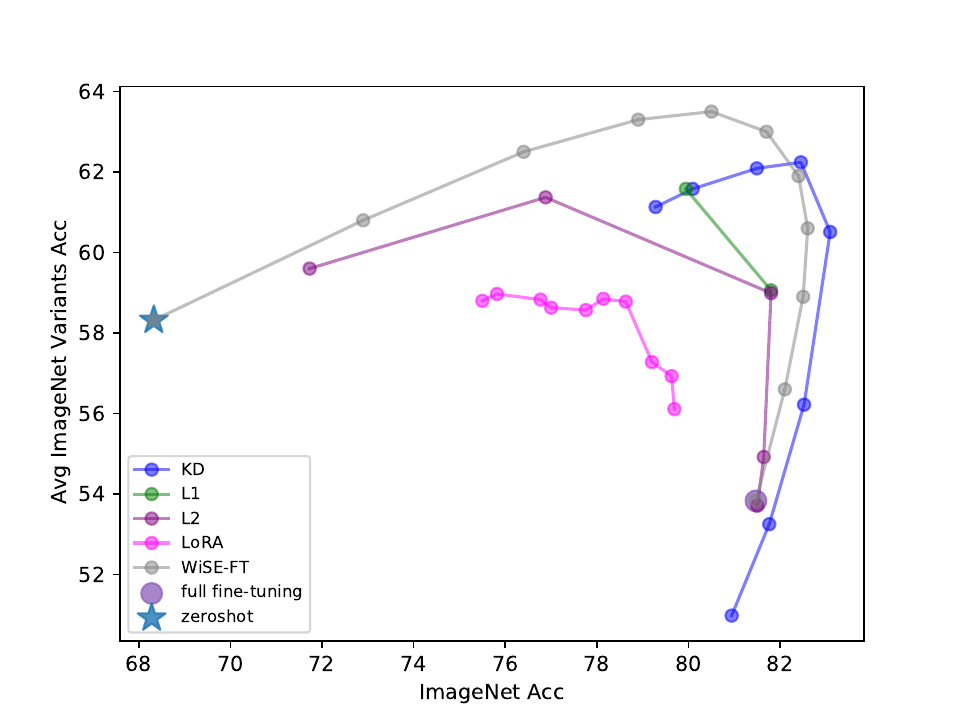}
    \includegraphics[width=0.45\linewidth]{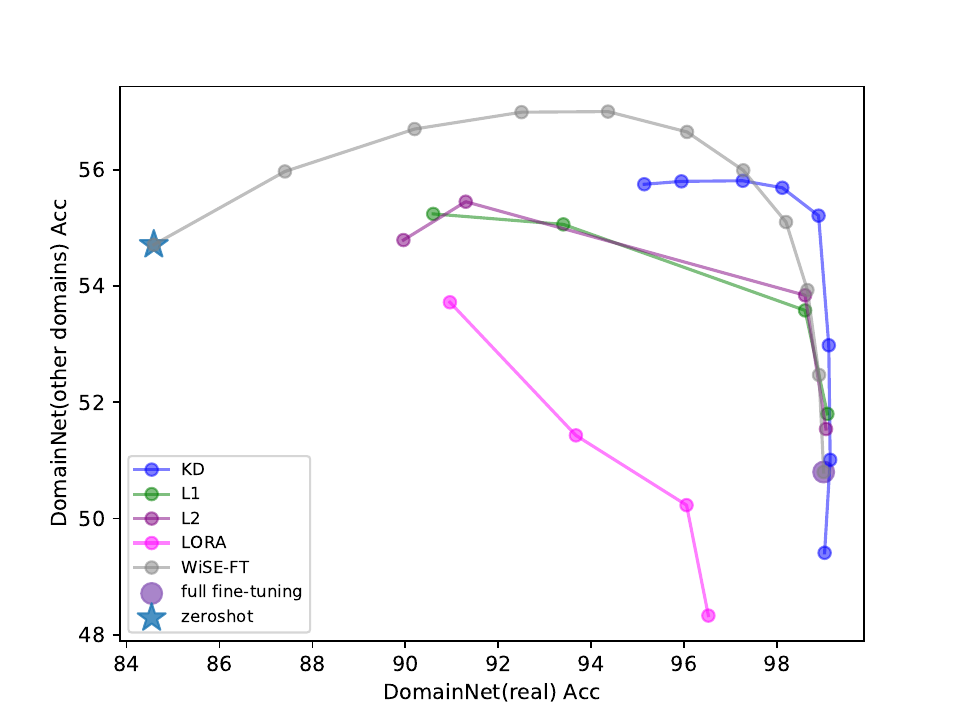}
    \caption{ID and OOD trade-off. (Left): fine-tune on ImageNet and evaluate the OOD performance by the average performance of ImageNet variants, i.e., ImageNet-V2, ImageNet-R, ImageNet-Sketch, ObjectNet; (Right) fine-tune on the `real' domain of DomainNet and evaluate the generality by the average performance on the Clipart”, “Infograph”, “Painting”, “Quickdraw”, and “Sketch” domains. Different points with the same color mean different parameters. }
    \label{fig:clip_imagenet_domainnet_results}
\end{figure}

The results of CLIP are presented in Figure~\ref{fig:clip_imagenet_domainnet_results}. Numerical results are shown in Appendix~\ref{app:detailed_results}. The left and right panels of Figure~\ref{fig:clip_imagenet_domainnet_results} showcase the outcomes of fine-tuning CLIP on ImageNet and DomainNet, respectively. 
It can be observed that \textbf{feature protection methods can boost OOD performance}. 

\begin{itemize}
\item L1 and L2 penalty both promote OOD performance, indicating that feature forgetting can be mitigated by regularizing fine-tuned weights towards pre-trained ones.
\item WiSE-FT stands out as it significantly alleviates feature forgetting and performs best. 
In ImageNet, WiSE-FT surpasses both the pre-trained and fine-tuned models, as well as methods like L1/L2/LoRA/KD, with the highest OOD performance exceeding 63\%. 
The trend observed in DomainNet is similar to that in ImageNet.
\item The KD method achieves better in-domain performance than WiSE-FT while maintaining a relatively high OOD performance. Specifically, KD and WiSE-FT are comparable to each other, and there is no consistent superiority of one over the other.
\item The trade-off of LoRA is inferior compared to other methods. We note that LoRA can not match the full fine-tuning performance on in-domain performance.
\end{itemize}

\section{Conclusion and Limitations}
In this work, we investigate why the OOD performance drops after fine-tuning. We find that the pre-trained model can lose semantic features and learn domain-specific features.
To protect pre-trained features, we explore various regularization methods from continual learning, as well as the weight averaging method (WiSE-FT) and parameter-efficient fine-tuning techniques like LoRA. Our findings demonstrate that continual learning and WiSE-FT methods effectively alleviate the forgetting of pre-trained features, with WiSE-FT outperforming others in achieving a balance between ID and OOD performance.

\bibliography{iclr2021_conference}

\begin{thebibliography}{25}
\providecommand{\natexlab}[1]{#1}
\providecommand{\url}[1]{\texttt{#1}}
\expandafter\ifx\csname urlstyle\endcsname\relax
  \providecommand{\doi}[1]{doi: #1}\else
  \providecommand{\doi}{doi: \begingroup \urlstyle{rm}\Url}\fi

\bibitem[Andreassen et~al.(2021)Andreassen, Bahri, Neyshabur, and Roelofs]{andreassen2021evolution}
Anders Andreassen, Yasaman Bahri, Behnam Neyshabur, and Rebecca Roelofs.
\newblock The evolution of out-of-distribution robustness throughout fine-tuning.
\newblock \emph{arXiv preprint arXiv:2106.15831}, 2021.

\bibitem[Barbu et~al.(2019)Barbu, Mayo, Alverio, Luo, Wang, Gutfreund, Tenenbaum, and Katz]{barbu2019objectnet}
Andrei Barbu, David Mayo, Julian Alverio, William Luo, Christopher Wang, Dan Gutfreund, Josh Tenenbaum, and Boris Katz.
\newblock Objectnet: A large-scale bias-controlled dataset for pushing the limits of object recognition models.
\newblock \emph{Advances in neural information processing systems}, 32, 2019.

\bibitem[Buzzega et~al.(2020)Buzzega, Boschini, Porrello, Abati, and Calderara]{buzzega2020dark}
Pietro Buzzega, Matteo Boschini, Angelo Porrello, Davide Abati, and Simone Calderara.
\newblock Dark experience for general continual learning: a strong, simple baseline.
\newblock \emph{Advances in neural information processing systems}, 33:\penalty0 15920--15930, 2020.

\bibitem[Davari et~al.(2022)Davari, Asadi, Mudur, Aljundi, and Belilovsky]{davari2022probing}
MohammadReza Davari, Nader Asadi, Sudhir Mudur, Rahaf Aljundi, and Eugene Belilovsky.
\newblock Probing representation forgetting in supervised and unsupervised continual learning.
\newblock In \emph{Proceedings of the IEEE/CVF Conference on Computer Vision and Pattern Recognition}, pp.\  16712--16721, 2022.

\bibitem[Goyal et~al.(2022)Goyal, Kumar, Garg, Kolter, and Raghunathan]{Goyal2022FLYP}
Sachin Goyal, Ananya Kumar, Sankalp Garg, Zico Kolter, and Aditi Raghunathan.
\newblock Finetune like you pretrain: Improved finetuning of zero-shot vision models.
\newblock \emph{ArXiv}, abs/2212.00638, 2022.
\newblock URL \url{https://api.semanticscholar.org/CorpusID:254125206}.

\bibitem[Hendrycks et~al.(2021{\natexlab{a}})Hendrycks, Basart, Mu, Kadavath, Wang, Dorundo, Desai, Zhu, Parajuli, Guo, et~al.]{hendrycks2021many}
Dan Hendrycks, Steven Basart, Norman Mu, Saurav Kadavath, Frank Wang, Evan Dorundo, Rahul Desai, Tyler Zhu, Samyak Parajuli, Mike Guo, et~al.
\newblock The many faces of robustness: A critical analysis of out-of-distribution generalization.
\newblock In \emph{Proceedings of the IEEE/CVF International Conference on Computer Vision}, pp.\  8340--8349, 2021{\natexlab{a}}.

\bibitem[Hendrycks et~al.(2021{\natexlab{b}})Hendrycks, Zhao, Basart, Steinhardt, and Song]{hendrycks2021natural}
Dan Hendrycks, Kevin Zhao, Steven Basart, Jacob Steinhardt, and Dawn Song.
\newblock Natural adversarial examples.
\newblock In \emph{Proceedings of the IEEE/CVF Conference on Computer Vision and Pattern Recognition}, pp.\  15262--15271, 2021{\natexlab{b}}.

\bibitem[Hu et~al.(2021{\natexlab{a}})Hu, Shen, Wallis, Allen-Zhu, Li, Wang, Wang, and Chen]{hu2021lora}
Edward~J Hu, Yelong Shen, Phillip Wallis, Zeyuan Allen-Zhu, Yuanzhi Li, Shean Wang, Lu~Wang, and Weizhu Chen.
\newblock Lora: Low-rank adaptation of large language models.
\newblock \emph{arXiv preprint arXiv:2106.09685}, 2021{\natexlab{a}}.

\bibitem[Hu et~al.(2021{\natexlab{b}})Hu, Shen, Wallis, Allen-Zhu, Li, Wang, and Chen]{Hu2021LoRALA}
J.~Edward Hu, Yelong Shen, Phillip Wallis, Zeyuan Allen-Zhu, Yuanzhi Li, Shean Wang, and Weizhu Chen.
\newblock Lora: Low-rank adaptation of large language models.
\newblock \emph{ArXiv}, abs/2106.09685, 2021{\natexlab{b}}.
\newblock URL \url{https://api.semanticscholar.org/CorpusID:235458009}.

\bibitem[Huang et~al.(2021)Huang, Zhang, Chen, Wang, and Yang]{huang2021continual}
Yufan Huang, Yanzhe Zhang, Jiaao Chen, Xuezhi Wang, and Diyi Yang.
\newblock Continual learning for text classification with information disentanglement based regularization.
\newblock \emph{arXiv preprint arXiv:2104.05489}, 2021.

\bibitem[Kirichenko et~al.(2022)Kirichenko, Izmailov, and Wilson]{kirichenko2022last}
Polina Kirichenko, Pavel Izmailov, and Andrew~Gordon Wilson.
\newblock Last layer re-training is sufficient for robustness to spurious correlations.
\newblock \emph{arXiv preprint arXiv:2204.02937}, 2022.

\bibitem[Kumar et~al.(2022)Kumar, Raghunathan, Jones, Ma, and Liang]{kumar2022fine}
Ananya Kumar, Aditi Raghunathan, Robbie Jones, Tengyu Ma, and Percy Liang.
\newblock Fine-tuning can distort pretrained features and underperform out-of-distribution.
\newblock \emph{arXiv preprint arXiv:2202.10054}, 2022.

\bibitem[Lin et~al.(2023)Lin, Yu, Kuang, Pathak, and Ramana]{Lin2023MultimodalityHU}
Zhiqiu Lin, Samuel Yu, Zhiyi Kuang, Deepak Pathak, and Deva Ramana.
\newblock Multimodality helps unimodality: Cross-modal few-shot learning with multimodal models.
\newblock \emph{ArXiv}, abs/2301.06267, 2023.
\newblock URL \url{https://api.semanticscholar.org/CorpusID:255942320}.

\bibitem[Panigrahi et~al.(2023)Panigrahi, Saunshi, Zhao, and Arora]{panigrahi2023task}
Abhishek Panigrahi, Nikunj Saunshi, Haoyu Zhao, and Sanjeev Arora.
\newblock Task-specific skill localization in fine-tuned language models.
\newblock \emph{arXiv preprint arXiv:2302.06600}, 2023.

\bibitem[Peng et~al.(2019)Peng, Bai, Xia, Huang, Saenko, and Wang]{peng2019moment}
Xingchao Peng, Qinxun Bai, Xide Xia, Zijun Huang, Kate Saenko, and Bo~Wang.
\newblock Moment matching for multi-source domain adaptation.
\newblock In \emph{Proceedings of the IEEE/CVF international conference on computer vision}, pp.\  1406--1415, 2019.

\bibitem[Qiu et~al.(2023)Qiu, Potapczynski, Izmailov, and Wilson]{qiu2023simple}
Shikai Qiu, Andres Potapczynski, Pavel Izmailov, and Andrew~Gordon Wilson.
\newblock Simple and fast group robustness by automatic feature reweighting.
\newblock \emph{arXiv preprint arXiv:2306.11074}, 2023.

\bibitem[Radford et~al.(2021)Radford, Kim, Hallacy, Ramesh, Goh, Agarwal, Sastry, Askell, Mishkin, Clark, et~al.]{radford2021learning}
Alec Radford, Jong~Wook Kim, Chris Hallacy, Aditya Ramesh, Gabriel Goh, Sandhini Agarwal, Girish Sastry, Amanda Askell, Pamela Mishkin, Jack Clark, et~al.
\newblock Learning transferable visual models from natural language supervision.
\newblock In \emph{International conference on machine learning}, pp.\  8748--8763. PMLR, 2021.

\bibitem[Recht et~al.(2019)Recht, Roelofs, Schmidt, and Shankar]{recht2019imagenet}
Benjamin Recht, Rebecca Roelofs, Ludwig Schmidt, and Vaishaal Shankar.
\newblock Do imagenet classifiers generalize to imagenet?
\newblock In \emph{International conference on machine learning}, pp.\  5389--5400. PMLR, 2019.

\bibitem[Rosenfeld et~al.(2022)Rosenfeld, Ravikumar, and Risteski]{rosenfeld2022domain}
Elan Rosenfeld, Pradeep Ravikumar, and Andrej Risteski.
\newblock Domain-adjusted regression or: Erm may already learn features sufficient for out-of-distribution generalization.
\newblock \emph{arXiv preprint arXiv:2202.06856}, 2022.

\bibitem[Tian et~al.(2023)Tian, Dai, Ma, He, Liu, and Kira]{Tian2023TrainablePG}
Junjiao Tian, Xiaoliang Dai, Chih-Yao Ma, Zecheng He, Yen-Cheng Liu, and Zsolt Kira.
\newblock Trainable projected gradient method for robust fine-tuning.
\newblock \emph{ArXiv}, abs/2303.10720, 2023.
\newblock URL \url{https://api.semanticscholar.org/CorpusID:257631710}.

\bibitem[Wang et~al.(2019)Wang, Ge, Lipton, and Xing]{wang2019learning}
Haohan Wang, Songwei Ge, Zachary Lipton, and Eric~P Xing.
\newblock Learning robust global representations by penalizing local predictive power.
\newblock \emph{Advances in Neural Information Processing Systems}, 32, 2019.

\bibitem[Wortsman et~al.(2021)Wortsman, Ilharco, Li, Kim, Hajishirzi, Farhadi, Namkoong, and Schmidt]{Wortsman2021wiseft}
Mitchell Wortsman, Gabriel Ilharco, Mike Li, Jong~Wook Kim, Hannaneh Hajishirzi, Ali Farhadi, Hongseok Namkoong, and Ludwig Schmidt.
\newblock Robust fine-tuning of zero-shot models.
\newblock \emph{2022 IEEE/CVF Conference on Computer Vision and Pattern Recognition (CVPR)}, pp.\  7949--7961, 2021.
\newblock URL \url{https://api.semanticscholar.org/CorpusID:237420687}.

\bibitem[Wortsman et~al.(2022)Wortsman, Ilharco, Kim, Li, Kornblith, Roelofs, Lopes, Hajishirzi, Farhadi, Namkoong, et~al.]{wortsman2022robust}
Mitchell Wortsman, Gabriel Ilharco, Jong~Wook Kim, Mike Li, Simon Kornblith, Rebecca Roelofs, Raphael~Gontijo Lopes, Hannaneh Hajishirzi, Ali Farhadi, Hongseok Namkoong, et~al.
\newblock Robust fine-tuning of zero-shot models.
\newblock In \emph{Proceedings of the IEEE/CVF Conference on Computer Vision and Pattern Recognition}, pp.\  7959--7971, 2022.

\bibitem[Xuhong et~al.(2018)Xuhong, Grandvalet, and Davoine]{xuhong2018explicit}
LI~Xuhong, Yves Grandvalet, and Franck Davoine.
\newblock Explicit inductive bias for transfer learning with convolutional networks.
\newblock In \emph{International Conference on Machine Learning}, pp.\  2825--2834. PMLR, 2018.

\bibitem[Zhao \& Yu(2006)Zhao and Yu]{zhao2006model}
Peng Zhao and Bin Yu.
\newblock On model selection consistency of lasso.
\newblock \emph{The Journal of Machine Learning Research}, 7:\penalty0 2541--2563, 2006.

\end{thebibliography}
\bibliographystyle{iclr2021_conference}

\appendix
\section{Accuracy and linear probing accuracy of 5 OOD domains in DomainNet.}
\label{app:all_target_domains}
\begin{figure}[H]
    \centering
    \includegraphics[width=0.8\linewidth]{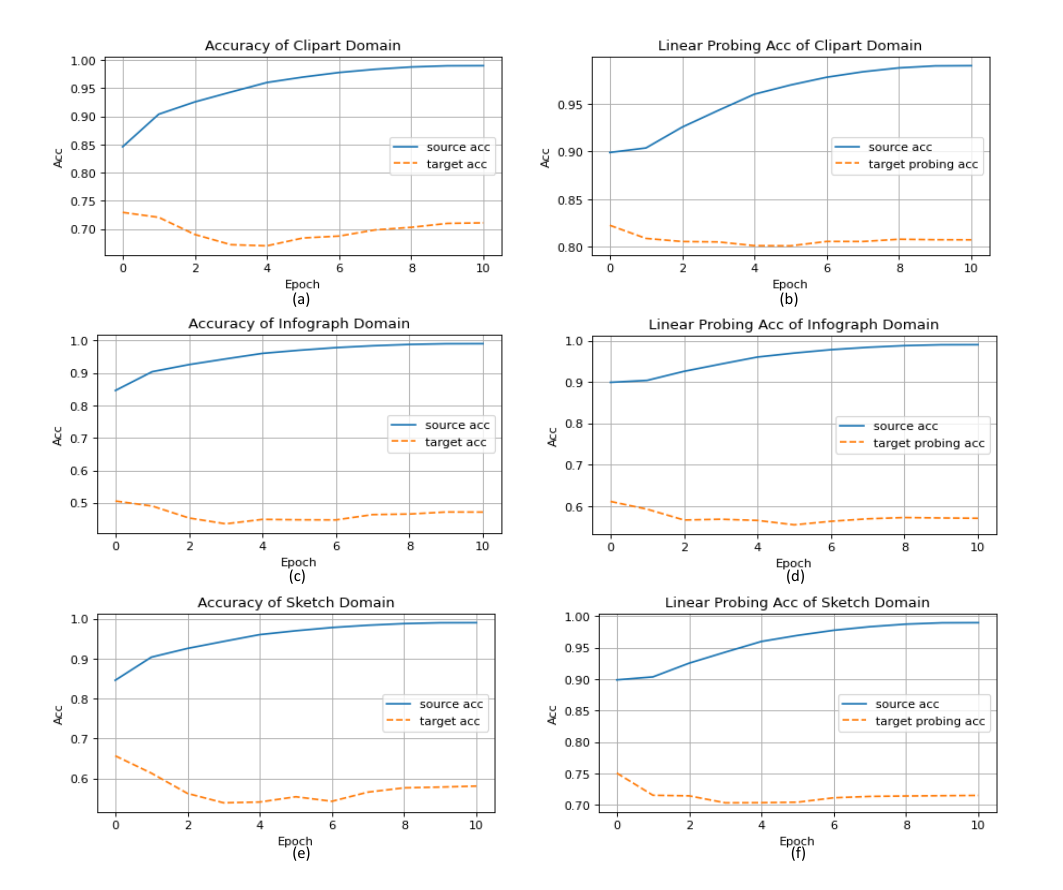}
    \caption{Accuracy and linear probing accuracy of 5 OOD domains in DomainNet.}
    \label{fig:all_target_domains}
\end{figure}

\section{Illustration Samples of Datasets}
\begin{figure}[H]
    \centering
    \includegraphics[width=0.45\linewidth]{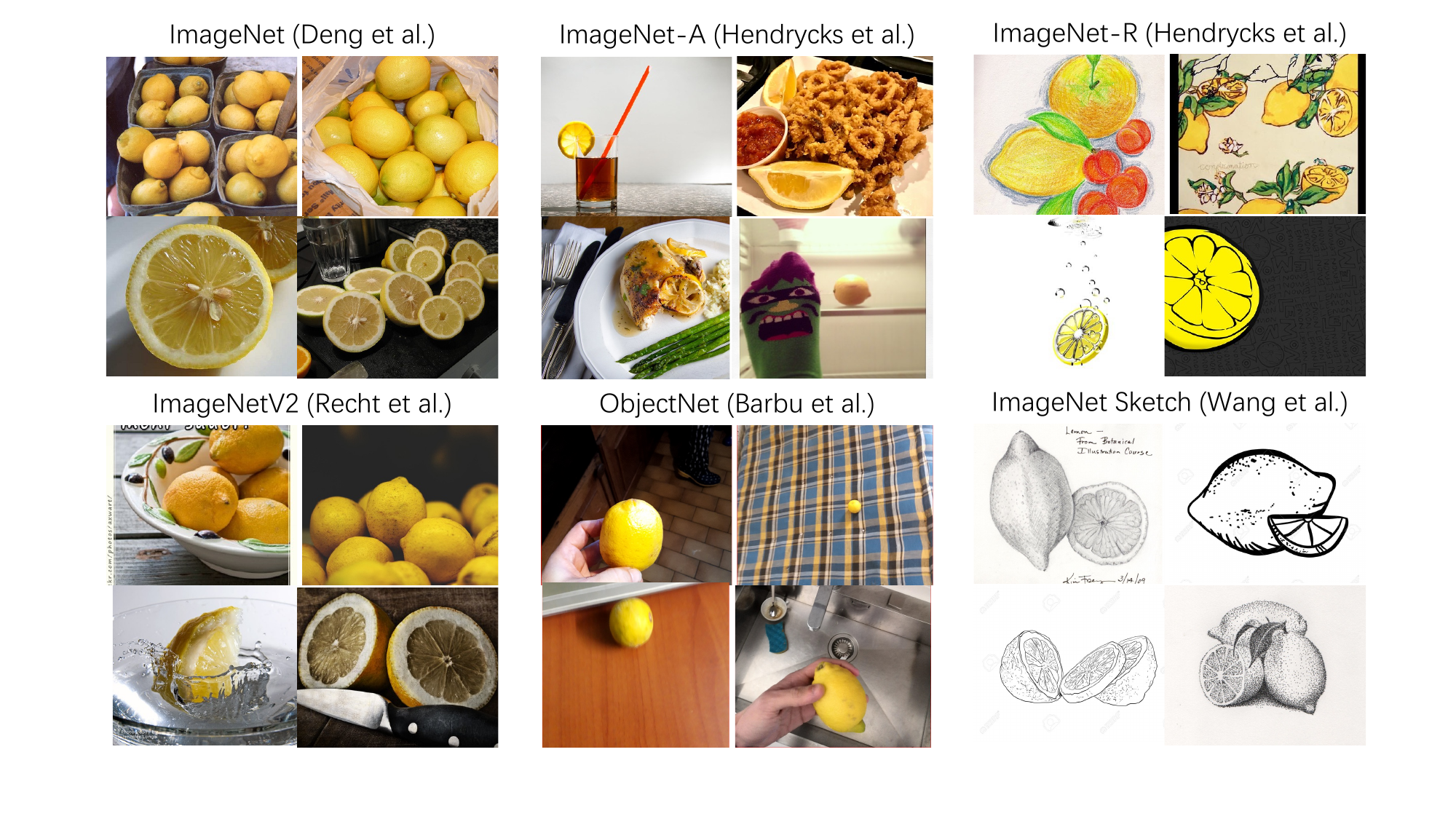}
    \includegraphics[width=0.45\linewidth]{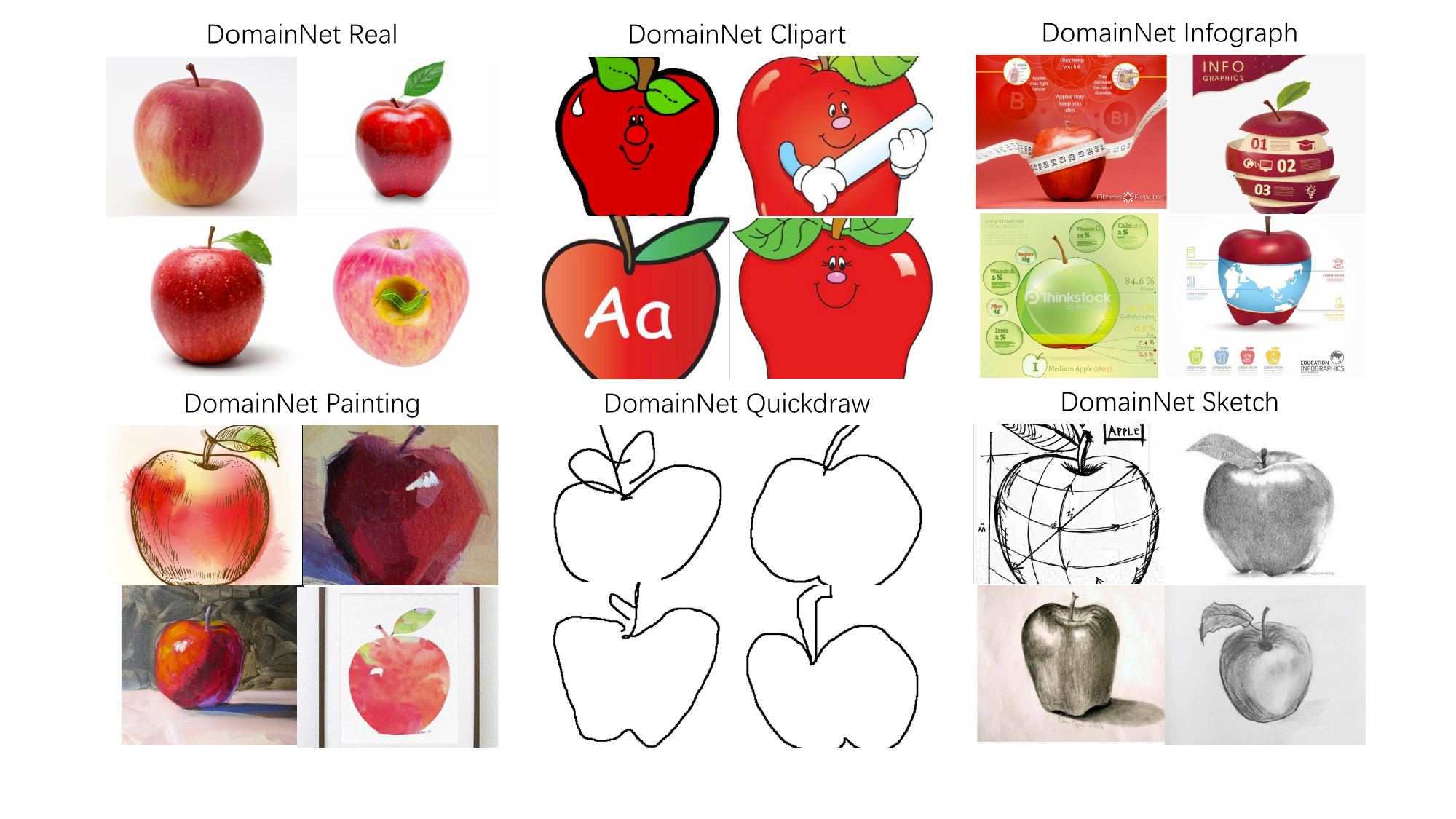}
    \caption{(Left) Illustration samples of the class lemon from ImageNet and 5 variants; (Right) Illustration samples of the class apple from DomainNet.}
    \label{fig:imagenet_samples}
\end{figure}

\section{Detailed Results}
\label{app:detailed_results}
\begin{table}[H]
    \caption{Accuracy(\%) results of ImageNet and derived distribution shifts after fine-tuning CLIP ViT-B/16 on ImageNet. }
    \centering
    \resizebox{\linewidth}{!}{
    \begin{tabular}{c|c|c|ccccc|c}
    \toprule
        Finetune Methods & Hyper & IN(ImageNet)&IN-V2& IN-R & IN-A &IN-Sketch& ObjectNet & Avg OOD \\ 
        \midrule
        ZeroShot && 68.33 & 62.00 & 77.62 & 49.96 & 48.26 & 53.77 & 58.32\\ 
        Full Fine-Tune && 81.47 & 70.89 & 65.34 & 36.92 & 45.83 & 50.18 & 53.83\\ \midrule
        \multirow{4}{*}{WiSE-FT} & $\alpha=0.2$ & 76.4 & 68.7 & 80.1 & 52.5 & 57.1 & 54.2 & 62.5\\ 
         &$\alpha=0.4$& 80.5 & 72.1 & 79.6 & 54.1 & 57.7 & 53.8 & 63.5\\ 
         &$\alpha=0.6$& 82.4 & 72.9 & 77.2 & 53.4 & 56.2 & 50.0 & 61.9 \\ 
         &$\alpha=0.8$& 82.5 & 72.8 & 72.7 & 51.0 & 53.5 & 44.6 & 58.9 \\ \midrule
        \multirow{2}{*}{l1} & $\lambda=1e-4$ & 79.94 & 71.10 & 77.23 & 52.15 & 51.30 & 56.12 & 61.58 \\ 
        & $\lambda=1e-5$ &81.80 & 72.26 & 72.24 & 46.32 & 50.26 & 54.23 & 59.06 \\ \midrule
        \multirow{5}{*}{l2}&$\lambda=1e-0$&71.73 & 64.35 & 78.48 & 50.54 & 49.34 & 55.29 & 59.60 \\
        &$\lambda=1e-1$& 76.88 & 69.10 & 78.47 & 52.68 & 50.45 & 56.16 & 61.37 \\ 
         &$\lambda=1e-2$&81.80 & 71.91 & 72.22 & 46.58 & 50.37 & 53.89 & 58.99 \\ 
         &$\lambda=1e-3$ & 81.64 & 71.54 & 66.91 & 38.65 & 46.89 & 50.61 & 54.92 \\ 
         &$\lambda=1e-4$ & 81.50 & 70.94 & 65.19 & 36.57 & 45.54 & 50.33 & 53.71 \\
         \midrule
        \multirow{4}{*}{KD} &$\lambda=5e-3$&81.49 & 72.46 & 78.65 & 50.43 & 52.18 & 56.73 & 62.09 \\ 
        &$\lambda=3e-3$& 82.45 & 73.32 & 78.5 & 50.27 & 52.2 & 56.92 & 62.24 \\
        &$\lambda=1e-3$& 83.09 & 73.57 & 76.03 & 46.77 & 51.05 & 55.11 & 60.51 \\
        &$\lambda=3e-4$& 82.52 & 72.77 & 70.29 & 39.16 & 47.47 & 51.39 & 56.22 \\ \midrule
        \multirow{5}{*}{LoRA} &\mbox{rank}=2& 75.82 & 68.13 & 74.86 & 49.18 & 48.15 & 54.54 & 58.97 \\ 
         &\mbox{rank}=4&77.00 & 68.88 & 73.49 & 48.37 & 47.90 & 54.52 & 58.63 \\ 
         &\mbox{rank}=8& 78.63 & 69.69 & 72.57 & 48.62 & 48.03 & 55.01 & 58.78 \\
         &\mbox{rank}=16& 79.20 & 70.12 & 70.47 & 45.14 & 47.34 & 53.34 & 57.28 \\
         &\mbox{rank}=32& 79.63 & 69.95 & 69.31 & 45.37 & 46.59 & 53.45 & 56.93\\  
        \bottomrule
    \end{tabular}
    }
    \label{tab:llm_med_medical}
\end{table}

\begin{table}[H]
    \caption{Accuracy(\%) results of DomainNet-real and other domains of DomainNet after fine-tuning CLIP ViT-B/16 on DomainNet-Real. }
    \centering
    \resizebox{\linewidth}{!}{
    \begin{tabular}{c|c|c|ccccc|c}
    \toprule
        Finetune Methods & Hyper & Real & Clipart& Infograph & Painting & Quickdraw & sketch & Avg OOD \\ 
        \midrule
        ZeroShot && 84.59 & 72.93 & 50.55 & 68.71 & 15.68 & 65.69 & 54.71\\ 
        Full Fine-Tune && 99.0 & 71.07 & 47.15 & 64.78 & 12.9 & 58.12 & 50.8\\ \midrule
        \multirow{4}{*}{WiSE-FT} & $\alpha=0.2$ & 90.2 & 75.47 & 52.91 & 70.83 & 16.97 & 67.31 & 56.7\\ 
         &$\alpha=0.4$& 94.36 & 76.22 & 53.56 & 71.05 & 17.16 & 67.0 & 57.0 \\ 
         &$\alpha=0.6$& 97.27 & 75.69 & 52.67 & 69.9 & 16.35 & 65.36 & 55.99 \\ 
         &$\alpha=0.8$& 98.65 & 73.99 & 50.54 & 67.85 & 14.87 & 62.37 & 53.93 \\ \midrule
        \multirow{4}{*}{l1} & $\lambda=1e-3$ &90.6 & 74.71 & 52.64 & 69.05 & 14.66 & 65.14 & 55.24 \\ 
        &$\lambda=1e-4$& 93.4 & 75.19 & 52.0 & 68.61 & 14.92 & 64.58 & 55.06 \\
        &$\lambda=1e-5$&93.4 & 75.19 & 52.0 & 68.61 & 14.92 & 64.58 & 55.06 \\
        &$\lambda=1e-6$&99.08 & 71.8 & 48.32 & 65.83 & 13.52 & 59.54 & 51.8 \\
        \midrule
        \multirow{4}{*}{l2}&$\lambda=1e-0$&89.96 & 73.61 & 52.77 & 68.56 & 14.14 & 64.85 & 54.79\\
        &$\lambda=1e-1$& 91.3 & 75.07 & 52.81 & 69.21 & 14.9 & 65.24 & 55.45 \\ 
         &$\lambda=1e-2$&91.3 & 75.07 & 52.81 & 69.21 & 14.9 & 65.24 & 55.45 \\ 
         &$\lambda=1e-3$ & 99.05 & 71.9 & 48.07 & 65.45 & 12.42 & 59.85 & 51.54 \\ 
         \midrule
        \multirow{4}{*}{KD} &$\lambda=1e-2$&95.14 & 75.39 & 52.15 & 69.73 & 15.49 & 66.01 & 55.75 \\ 
        &$\lambda=1e-3$&  98.89 & 75.46 & 51.8 & 69.23 & 14.65 & 64.9 & 55.21 \\
        &$\lambda=1e-4$& 99.14 & 71.59 & 47.06 & 64.66 & 12.65 & 59.11 & 51.01 \\
        &$\lambda=1e-5$& 99.02 & 69.56 & 44.63 & 63.33 & 12.65 & 56.86 & 49.41 \\ \midrule
        \multirow{1}{*}{LoRA} &\mbox{rank}=16& 96.05 & 69.94 & 46.02 & 64.26 & 12.64 & 58.29 & 50.23  \\
        \bottomrule
    \end{tabular}
    }
    \label{tab:llm_med_medical}
\end{table}

\section{Method Details}
\label{app:LoRA_details}
\textbf{LoRA.}
In our study, we apply LoRA specifically to two weight matrices ($W_q$ and $W_v$) within the self-attention module of the Transformer architecture. We constrain the update of a pre-trained weight matrix $\Delta \theta_0 = \theta - \theta_0 \in \mathbb{R}^{d \times k}$ by representing the updated portion as $\Delta\theta = BA$, where $B \in \mathbb{R}^{d \times r}$, $A \in \mathbb{R}^{r \times k}$, and the rank $r$ is much smaller than $min(d,k)$. We explore different values of $r$ as a hyper-parameter. During training, $\theta_0$ remains fixed, and only $A$ and $B$ receive gradient updates. We initialize $A$ with random Gaussian values and set $B$ to zero.

\section{Experimental Details}
\label{app:exp_details}
\begin{table}[H]
    \caption{Details of fine-tuning CLIP}
    \centering
    \begin{tabular}{c|c|c}
    \toprule
        Training Dataset & ImageNet & DomainNet(`Real') \\
        \midrule
        Optimizer & AdamW & AdamW \\
        Optimizer Hyper-parameter & warmup 500 steps & warmup 500 steps \\
        Learning Rate Schedule & Cosine & Cosine \\
        Warmup & 500 Steps & 500 steps\\
        Learning Rate & 3e-5 & 3e-5 \\
        Epoch & 10 & 10 \\
        Steps per epoch & 2502 & 342 \\
        Batch Size & 512 & 512\\
        \bottomrule
    \end{tabular}
    \label{tab:clip_details}
\end{table}

We use the CLIP model ViT-B/16\cite{radford2021learning}. We fine-tune the pre-trained model on ImageNet and DomainNet-Real. We use the AdamW optimizer with the default PyTorch settings and choose 512 as batch size. We use a learning rate of $3 \times 10^{-5}$, gradient clipping at global norm 1, and fine-tune for a total of 10 epochs. The settings mentioned above are the same with \cite{wortsman2022robust}.

\section{Limitations}
One limitation is that we haven't covered the rehearsal methods, which means replaying a small portion of the pre-trained dataset. 
Another limitation is that we have not explored the impact of varying model sizes on the forgetting issue and the corresponding methods. We plan to investigate these aspects in future work.

\end{document}